\documentclass[a4paper,fleqn]{cas-dc}

\usepackage[authoryear]{natbib}
\usepackage{array}

\usepackage{enumitem}
\usepackage{xcolor}
\usepackage{booktabs} 
\usepackage{multirow} 
\usepackage{siunitx} 
\usepackage{amssymb}  

\def\tsc#1{\csdef{#1}{\textsc{\lowercase{#1}}\xspace}}
\tsc{WGM}
\tsc{QE}
\tsc{EP}
\tsc{PMS}
\tsc{BEC}
\tsc{DE}

\ExplSyntaxOn
\cs_set:Npn \__first_footerline:
  {
    \group_begin:
    \small\sffamily
    \ifnum\theblind>0\relax
    \else
      \__short_authors:
    \fi
    \group_end:
  }
\ExplSyntaxOff

\begin{document}
\let\WriteBookmarks\relax
\def\floatpagepagefraction{1}
\def\textpagefraction{.001}

\shorttitle{Impact of
interpulse\--interval variation on bat classifiers}

\shortauthors{W. Eversteijn et~al.}

\title [mode = title]{Effects of interpulse-interval variation on deep-learning classification of bat vocalizations}      

\author[1,2]{Welmoed Eversteijn}[orcid=0009-0003-0036-025X]
\ead{welmoed.eversteijn@gmail.com}
\credit{Conceptualization, Methodology, Software, Validation, Formal analysis, Investigation, Data Curation, Writing - Original Draft, Writing - Review \& Editing, Visualization}

\author[1]{Burooj Ghani}[orcid=0000-0002-0541-3571]
\ead{burooj.ghani@naturalis.nl}
\credit{Conceptualization, Methodology, Software, Resources, Writing - Review \& Editing, Supervision}

\author[1]{A. Leonie Baier}[orcid=0000-0002-0327-0378]
\cormark[1] 
\ead{leonie.baier@naturalis.nl}
\credit{Conceptualization, Resources, Data Curation, Writing - Original Draft, Writing - Review \& Editing, Visualization, Supervision}

\author[1,2]{Dan Stowell}[orcid=0000-0001-8068-3769]
\ead{dan.stowell@naturalis.nl}
\credit{Conceptualization, Resources, Writing - Review \& Editing, Supervision, Project administration, Funding acquisition}

\affiliation[1]{organization={Naturalis Biodiversity Centre},
    addressline={Darwinweg 2},
    city={Leiden},
    postcode={2333 CR},
    country={The Netherlands}}

\affiliation[2]{organization={Department of Cognitive Science and Artificial Intelligence, Tilburg University},
    addressline={Warandelaan 2},
    city={Tilburg},
    postcode={5037 AB},
    country={The Netherlands}}

\cortext[cor1]{Corresponding author}

\begin{abstract}
Temporal context may aid automated bat-species classification, but the contribution of specific features remains unclear. We investigated whether variation in the interpulse interval (IPI)-the time between consecutive call onsets-provides species-discriminative information and whether transformer-based models are more sensitive to this information than convolutional neural networks. We created two matched datasets from European bat recordings: a natural-IPI condition retaining the original call timing and a normalized-IPI condition in which call onsets were spaced at 50-ms intervals. EfficientNet-B0 and PaSST were fine-tuned and evaluated within each condition. In an additional experiment, each architecture was trained separately on natural-IPI and normalized-IPI recordings, and evaluated on the same natural-IPI test set. Finally, the pretrained classifiers BatDetect2 and BAT were evaluated on both conditions. Within-condition IPI normalization had model-dependent effects. PaSST accuracy differed little between the natural-IPI (\(71 \pm 2.3\%\)) and normalized-IPI (\(70 \pm 6.3\%\)) conditions, whereas EfficientNet accuracy increased from \(47 \pm 4.7\%\) to \(57 \pm 3.9\%\). PaSST exceeded EfficientNet under both conditions. In the cross-condition evaluation, models trained on natural-IPI recordings outperformed those trained on normalized-IPI recordings on the natural-IPI test set: accuracy decreased from 54\% to 50\% for EfficientNet and from 65\% to 57\% for PaSST. BatDetect2 and BAT differed little between IPI conditions. Overall, we found limited support for the hypotheses that natural IPI variation contributes substantially to bat-species classification and that it is used more effectively by transformer-based than CNN-based models. Nevertheless, the cross-condition performance decrease shows that results obtained under normalized conditions may not transfer fully to natural recordings.
\end{abstract}


\begin{keywords}
Bioacoustics \sep Deep learning \sep Temporal dynamics \sep Automated species identification 
\end{keywords}

\maketitle

\section{Introduction}
Biodiversity is rapidly decreasing on a global scale, highlighting the need for large-scale ecological monitoring methods that are affordable. Monitoring species populations helps assess the health of the ecosystem and evaluate conservation efforts. Bats specifically, are effective as bioindicators because they are sensitive to environmental changes reflecting ecosystem shifts~\citep{russoWeNeedUse2021}. Additionally, bats provide essential ecosystem services, including insect pest control, pollination and seed dispersal~\citep{ancillotto2017Effects, ghanemIncreasingAwarenessEcosystem2012}. However, observing the nocturnal animals during the night is a difficult task, making acoustic monitoring a well-established tool in the study of bats. In the past, handheld detectors were used to help identify bats during field observations. Unfortunately, these only provide snapshot representations of the species distributions and do not accurately portray the biodiversity.  
Passive acoustic monitoring (PAM) advances this approach by applying autonomous acoustic sensors that record environmental sounds continuously. Recently, these sensors have become less expensive, enabling large sensor networks and making large-scale monitoring more feasible~\citep{kobayashi2021Development}. 
Finally, to address the growing increase of acoustic data, automated bat call classification systems have been developed.  Machine learning models have shown promise. Most existing studies employ convolutional neural networks (CNNs), such as BatNet, NABat ML and BatDetect2~\citep{chen2020AutomaticStandardizedProcessing,khalighifar2022NABat,macaodhaGeneralApproachBat2022}. These learn local spatial patterns from spectrogram representations of bat calls. More recently, transformer architectures have gained attention in audio classification because the self-attention mechanisms allow interactions between further elements within an input sequence, potentially enabling the use of broader temporal context information~\citep{koutini2022Learning}.

\begin{figure*}
	\centering
		\includegraphics[width=\textwidth]{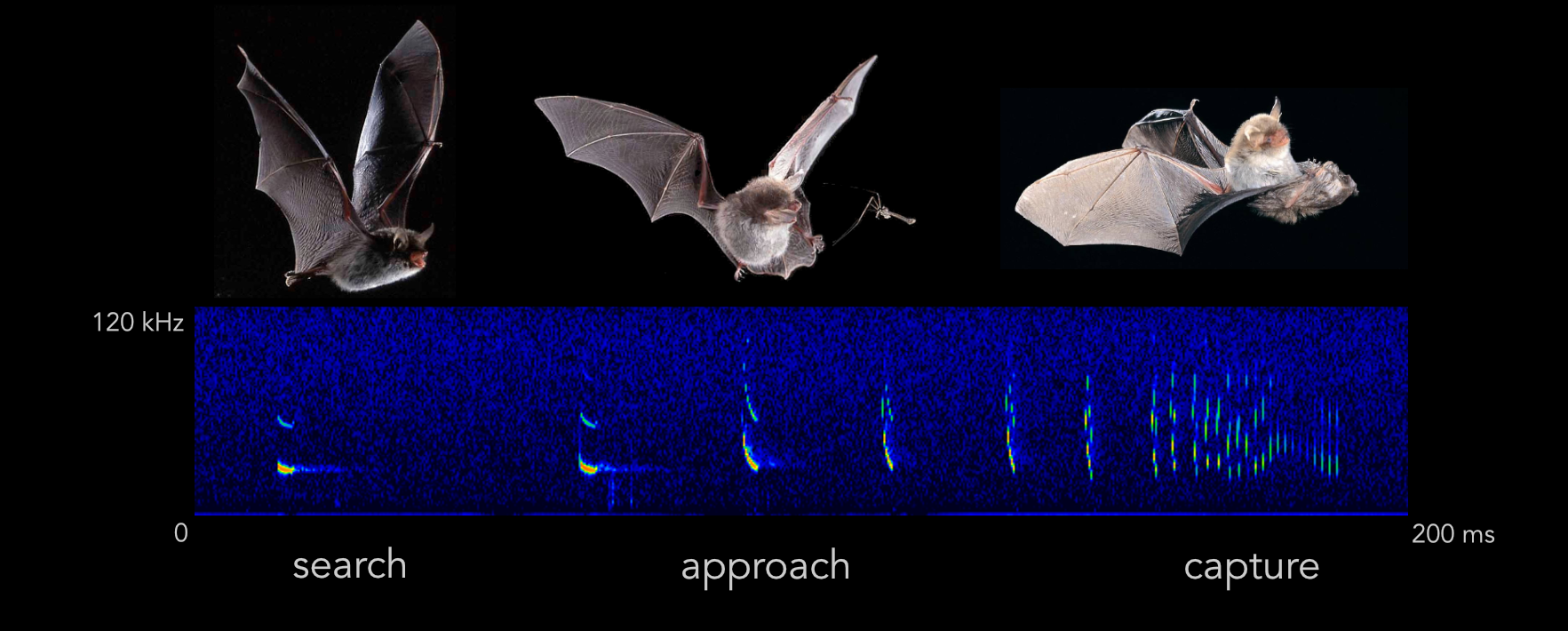}
	\caption{Spectrogram of a bat echolocation sequence illustrating the temporal progression from search to approach and capture (terminal buzz), and the corresponding changes in call structure. Approximate frequency and time scales are included to provide a general sense of scale. The photographs illustrate the three behavioural phases; the depicted bat species differs from that of the echolocation sequence. Bat photographs: © Dietmar Nill, adapted and used with permission. The photographs are excluded from the Creative Commons license applicable to this article.}
	\label{fig:callsequence}
\end{figure*}

The morphological features of bat echolocation calls—such as duration, bandwidth, and frequency modulation steepness—are highly plastic, varying not only by species identity but also across geographical habitats and dynamic behavioral contexts~\citep{denzingerBatGuildsConcept2013}. During a hunting or navigation sequence, bats systematically alter their call structure as they transition from search to approach and capture phases, typically producing shorter, more broadband signals as they close in on the target (Figure~\ref{fig:callsequence}). This intra-species variation introduces significant overlap in the feature space of different species. For instance, a highly broadband call could represent a \textit{Myotis nattereri} early in its search phase, or a \textit{Myotis daubentonii} late in its approach phase. Theoretically, automated classifiers could resolve this ambiguity if they possess the capacity to contextualize an individual call within the broader temporal sequence. By leveraging the surrounding temporal structure—specifically the interpulse interval (IPI) variation that demarcates these behavioral transitions—a model could infer a call's relative position in the sequence, using this contextual anchor to improve species discrimination.

Here, we investigate whether temporal information in the form of interpulse interval (IPI) variation contributes to automated bat-species classification, and whether transformer-based architectures leverage this sequence context more effectively than traditional CNNs. To isolate this effect, we generated matched datasets where the natural temporal spacing between consecutive calls was either preserved or normalized to a fixed interval while preserving the spectral content of individual pulses. Through this controlled manipulation of call timing, we peer inside the 'black box' of automated bat monitoring, revealing which acoustic features drive deep learning decisions and how to leverage them for more reliable wildlife surveys.

\section{Data and methods}
\subsection{Study design and objectives}
To evaluate the effect of sequence context on classification performance, we conducted three experiments (Table~\ref{tab:experimental-overview}) using two matched dataset conditions: a \textit{natural IPI} condition that retained the original temporal variation, and a \textit{normalized IPI} condition in which intervals between call onsets were normalized to 50~ms. 

The first experiment compared two pretrained modeling pipelines, taking the initial weights from EfficientNet-B0 and PaSST, pretrained on Audioset and ImageNet, respectively. All layers of each model were fine-tuned and the models were evaluated separately within each IPI condition using five-fold stratified cross-validation. Second, in a cross-condition, held-out experiment, each model was trained separately on natural-IPI and normalized-IPI recordings using a single train--test split. All four trained models were evaluated on the same fixed natural-IPI test set. Partitioning was performed at the source-recording level to ensure that matched versions of the same recording remained in the same partition.This experiment assessed whether training without natural temporal variation affected generalization to realistic recordings. In the third experiment, the existing bat classifiers BatDetect2 and BAT were evaluated on both dataset conditions to assess their inherent sensitivity to IPI variation.

\begin{table*}[width=\textwidth,cols=4,pos=t]
\caption{Overview of the experimental design. Natural IPI refers to recordings retaining the original temporal spacing between calls; normalized IPI refers to recordings in which variation in the intervals between call onsets was reduced.}
\label{tab:experimental-overview}

\begin{tabular*}{\tblwidth}{
@{}
>{\raggedright\arraybackslash}p{0.24\textwidth}
>{\raggedright\arraybackslash}p{0.11\textwidth}
>{\raggedright\arraybackslash}p{0.22\textwidth}
>{\raggedright\arraybackslash}p{0.31\textwidth}
@{}
}
\toprule
Experiment and purpose
& Models
& Training condition
& Evaluation procedure \\
\midrule

\textbf{1. Within-condition evaluation}\newline
Assess the effect of IPI normalization when training and evaluation use the same IPI condition
&
EfficientNet-B0\newline
PaSST
&
Separate training on:\newline
(1) natural IPI;\newline
(2) normalized IPI
&
Five-fold stratified cross-validation was conducted within each condition:\newline
natural IPI $\rightarrow$ natural IPI;\newline
normalized IPI $\rightarrow$ normalized IPI. \\

\addlinespace

\textbf{2. Cross-condition evaluation}\newline
Assess generalization from normalized to natural IPI
&
EfficientNet-B0\newline
PaSST
&
Separate training on:\newline
(1) natural IPI;\newline
(2) normalized IPI
&
Models were evaluated on natural-IPI test recordings:\newline
natural IPI $\rightarrow$ natural IPI (baseline);
\newline normalized IPI $\rightarrow$ natural IPI.

\\
\addlinespace

\textbf{3. Existing-model evaluation}\newline
Assess whether IPI normalization affects predictions from existing classifiers
&
BatDetect2\newline
BAT
&
No additional training;\newline
pretrained models were used
&
Models were evaluated on natural-IPI and matched normalized-IPI recordings.\newline
Performance was calculated for all 16 species and for only the species represented in each model's original training data. \\

\bottomrule
\end{tabular*}
\end{table*}

\subsection{Data collection and dataset composition}
To investigate the role of IPI variation, two matched dataset conditions were created from a subset of the European bat vocalization dataset ChiroSetEurope~\citep{baier2026CSEpaper}. The natural-IPI condition retained the original temporal spacing between consecutive calls, whereas in the normalized-IPI condition, intervals between call onsets were normalized using a custom annotation tool. The two conditions were intended to differ only in their temporal call structure.

The study aimed to include approximately 20 two-second recordings per species. The final number varied from 7 to 21 recordings per species, with over 15 available for most species (Table~\ref{tab:species_recordings}). Recordings were inspected manually in Audacity (version 3.4.2; Audacity Team, 2025), and two-second segments were selected based on the presence of multiple calls. The selected segments were exported as WAV files. Recordings were excluded if they contained overlapping species, had uncertain species labels, or contained too few calls. A duration of two seconds generally preserved a sequence of consecutive calls—necessary for examining IPI variation—while remaining computationally manageable.

Species selection prioritized the inclusion of as many members of the genus \textit{Myotis} as possible because these species are difficult to distinguish from their spectral call characteristics and frequently present challenges for automated classifiers~\citep{walters2012Continentalscale, bergmann2022Soundscape}. Additional species were included when ChiroSetEurope contained enough recordings that met the selection criteria.

Recordings in the source dataset varied in sampling rate, and some had undergone time expansion, which increases signal duration while proportionally lowering its frequencies. Consequently, a two-second segment from a time-expanded recording generally contained fewer calls than a segment of the same duration from the original recording. Because the study required sequences containing multiple calls, recordings with sampling rates below 100 kHz were excluded. Inspection of the source recordings in Audacity indicated that recordings sampled above this threshold generally contained non-time-expanded ultrasonic audio.

During the original analysis, recordings labelled as \textit{Cnephaeus serotinus} and \textit{Eptesicus serotinus} were encoded as separate classes. Subsequent taxonomic review established that these names refer to the same species, with \textit{E. serotinus} representing the former taxonomic combination of \textit{C. serotinus}~\citep{claudio2023taxonomic}. This nomenclatural duplication was identified after completion of model training and evaluation; consequently, the reported analyses retain the original class structure. The number of operational class labels therefore exceeds the number of biological species by one.

\begin{table*}[width=\textwidth,cols=5,pos=t]
\centering
\caption{Number and total duration of recordings per species. Checkmarks
indicate that the species was represented in the original training data of
BatDetect2 or BAT.}
\label{tab:species_recordings}

\begin{tabular*}{\tblwidth}{@{\extracolsep{\fill}} lrccc@{}}
\toprule
\textbf{Species}
& \makecell{\textbf{Number of}\\\textbf{recordings}}
& \makecell{\textbf{Total audio}\\\textbf{duration (sec)}}
& \makecell{\textbf{BatDetect2}\\\textbf{training set}}
& \makecell{\textbf{BAT}\\\textbf{training set}} \\
\midrule

\textit{Miniopterus schreibersii} & 21 & 42 & --         & \checkmark \\
\textit{Barbastella barbastellus}  & 20 & 40 & \checkmark & -- \\
\textit{Eptesicus serotinus}       & 20 & 40 & \checkmark & \checkmark \\
\textit{Hypsugo savii}             & 20 & 40 & --         & \checkmark \\
\textit{Myotis capaccinii}         & 20 & 40 & --         & \checkmark \\
\textit{Myotis dasycneme}          & 20 & 40 & --         & \checkmark \\
\textit{Myotis nattereri}          & 20 & 40 & \checkmark & \checkmark \\
\textit{Myotis brandtii}           & 18 & 36 & \checkmark & \checkmark \\
\textit{Myotis davidii}            & 16 & 32 & --         & -- \\
\textit{Myotis myotis}             & 16 & 32 & --         & \checkmark \\
\textit{Myotis mystacinus}         & 16 & 32 & \checkmark & \checkmark \\
\textit{Cnephaeus serotinus}       & 11 & 22 & \checkmark & \checkmark \\
\textit{Myotis emarginatus}        & 10 & 20 & --         & \checkmark \\
\textit{Myotis alcathoe}           & 8  & 16 & \checkmark & -- \\
\textit{Myotis bechsteinii}        & 8  & 16 & \checkmark & -- \\
\textit{Myotis blythii}            & 7  & 14 & --         & -- \\

\bottomrule
\end{tabular*}
\end{table*}

\subsection{IPI annotation and normalization}
To create the normalized-IPI dataset, each bat call was manually annotated in Audacity (version 3.4.2; Audacity Team, 2025) using a custom annotation tool (see Section \textit{Code availability}) connected through Audacity’s mod-script-pipe interface. The tool loaded each WAV file, verified its sampling rate, and extracted the selected two-second segment. For each call, an onset label marked the estimated beginning of the call, and an interval label encompassed its spectral energy, including visible echoes and harmonics. The interval labels identified audio that the preprocessing algorithm should treat as call content rather than background. Once annotation was complete, the labels and their timestamps were transferred to the preprocessing pipeline. IPI was defined as the time between the annotated onsets of two consecutive calls.

The target IPI was set to 50 ms because this interval exceeds the expected pulse durations of the included species, thereby reducing the risk of overlap between consecutive calls after normalization. The threshold was informed by North American bat literature~\citep{khalighifar2022NABat}, as no equivalent guideline for European species was identified.

For each pair of consecutive calls, the observed onset interval was compared with the target interval of \(0.050 \times f_s\) samples, where \(f_s\) is the sampling rate. Intervals exceeding the target were shortened by removing intervening audio, with background samples removed before any call-labelled samples were altered. Intervals shorter than the target were extended by inserting background audio from the same recording while preserving the original audio between the calls. The algorithm preferentially used a single continuous background segment; when no segment of sufficient duration was available, multiple smaller segments were combined.

Following these modifications, call onsets were aligned to a regular 50-ms sequence anchored at the first annotated onset. The target onset of call \(i\) was therefore:

\begin{equation}
    t_i = t_1 + (i-1)\times 0.050\,\mathrm{s},
    \label{eq:target-onset}
\end{equation}

where \(t_1\) is the onset of the first call.

The term “IPI normalization” refers to reducing variation in the intervals between consecutive calls. Although annotated onsets were placed on a regular 50-ms sequence, small deviations in the true acoustic onset positions may remain because of uncertainty in manual onset annotation.

\subsection{Model preprocessing, configuration and implementation}
Two pretrained models were fine-tuned: the CNN-based EfficientNet-B0, pretrained on ImageNet and implemented using timm version 1.0.24~\citep{rw2019timm}, and the transformer-based PaSST, pretrained on AudioSet~\citep{koutini2022Learning}. All model parameters were updated during fine-tuning. The complete model, preprocessing, and training configurations are reported in Table~\ref{tab:model_config}.

\begin{table}[ht]
\centering
\caption{Model, preprocessing, and training configurations.}
\label{tab:model_config}

\resizebox{\tblwidth}{!}{
\begin{tabular}{lll}
\toprule
\textbf{Parameter} & \textbf{EfficientNet} & \textbf{PaSST} \\
\midrule
Backbone & CNN & Transformer \\
Model version & EfficientNet-B0 & \makecell{passt\_s\_f128\_20sec\_p16\\\_s10\_ap.474\_swa.pt}\\
Pretraining & ImageNet & AudioSet \\
Input size & $3 \times 128 \times 1024$ & $1 \times 128 \times 2000$ \\
Clip duration & 2 sec & 20 sec \\
Sampling rate & 250 kHz & 32 kHz \\
$n_{\mathrm{fft}}$ & 2048 & 1024 \\
Window length & 2048 & 1024 \\
Hop size & 256 & 320 \\
Spectral power & 0.2 & 2.0 \\
Number of Mel bins & -- & 128 \\
Number of FFT bins & 1025 & 513 \\
Image height & 128 pixels & 128 bins \\
Image width & 1024 & 2000 \\
Input channels & 3 & 1 \\
Spectrogram type & Linear spectrogram & Mel spectrogram \\
Physical batch size & 4 & 1 \\
Gradient-accumulation steps & 8 & 32 \\
Effective batch size & 32 & 32 \\
Maximum epochs & 50 & 50 \\
Loss function & Cross-entropy & Cross-entropy \\
Optimizer & AdamW & AdamW \\
Learning-rate candidates
& 0.00001, 0.00002, 0.00005
& 0.00001, 0.00002, 0.00005 \\
Weight-decay candidates
& 0.00001, 0.0001, 0.001 & 0.00001, 0.0001, 0.001 \\
Early-stopping criterion
& Macro-averaged F1-score
& Macro-averaged F1-score \\
Early-stopping patience & 5 & 5 \\
\bottomrule
\end{tabular}
}
\end{table}

For EfficientNet-B0, the two-second audio recordings were resampled and converted into linear spectrograms. The spectrogram parameters were selected through visual inspection to ensure that call structure remained clearly represented. The resulting frequency bins were normalized using z-score normalization, and the spectrograms were resized using bilinear interpolation. Because EfficientNet requires three-channel input, each single-channel spectrogram was duplicated across three channels.

For PaSST, the recordings were time-expanded by a factor of 10 before being resampled to the input sampling rate required by the model. Time expansion reduced both call frequencies and playback speed by the same factor, shifting the approximate call-frequency range from 20–120 kHz to 2–12 kHz. Performing time expansion before resampling preserved this frequency-shifted call content; resampling the original recordings directly would instead have removed frequencies above the Nyquist limit of the target sampling rate. The PaSST preprocessing pipeline subsequently converted the waveforms into mel spectrograms.

Different physical batch sizes and gradient-accumulation settings were used to accommodate the memory requirements of the two models, while maintaining the same effective batch size. Both models were trained for a maximum of 50 epochs using early stopping based on validation macro-averaged F1-score.

The data and model pipelines were implemented in Python 3.8.20 using PyTorch 2.4.1, TorchAudio 2.4 and Pytorch Lightning. Model training was conducted on an in-house remote server with CUDA 12.2 equipped with two NVIDIA A40 GPUs with 46 GB of VRAM each.

\subsection{Experiment 1: within-condition training and cross-validation}

EfficientNet-B0 and PaSST were separately fine-tuned and evaluated within each IPI condition, resulting in four model–condition combinations. Performance was assessed using five-fold stratified cross-validation, with approximately 80\% of the recordings used for training and 20\% for validation in each fold. Identical fold assignments were used across models and IPI conditions. A weighted random sampler was applied to each training fold to account for the remaining class imbalance.
Model-specific preprocessing is described in Section 2.4, and the training configurations are reported in Table 3. Models were optimized using cross-entropy loss and AdamW. Hyperparameters were selected through grid search based on mean validation macro-averaged F1-score across the five folds. Early stopping was also based on validation macro-averaged F1-score. No independent test set was used in this experiment.

\subsection{Experiment 2: cross-condition held-out evaluation}

This experiment assessed whether training on normalized-IPI recordings affected generalization to recordings retaining natural temporal variation. EfficientNet-B0 and PaSST were each trained separately on the natural-IPI and normalized-IPI versions of the training recordings, resulting in four model runs. All four trained models were evaluated on the same fixed natural-IPI test set. Models trained and tested on natural-IPI recordings provided the matched-condition baseline, whereas models trained on normalized-IPI recordings and tested on natural-IPI recordings represented the cross-condition evaluation.

Unlike Experiment 1, which used five-fold cross-validation to estimate within-condition performance across multiple data partitions, Experiment 2 used a single stratified 80/20 train--test split. This design provided a fixed natural-IPI test set for directly comparing the two training conditions. The test recordings and their matched normalized-IPI versions were held out from model training across all runs, rather than rotating between the training and validation partitions as they did during cross-validation. The split contained approximately 16 training recordings and four test recordings per species. Partitioning was performed at the source-recording level using identifiers shared between the natural-IPI and normalized-IPI datasets. Both versions of each source recording were assigned to the same partition, preventing matched versions of the same recording from occurring across the training and test sets. Consequently, Experiment 2 produced a single accuracy estimate for each model and training condition rather than a mean and variability estimate across folds.

The model-specific preprocessing and training procedures followed those described previously. No separate validation set was created because the limited number of recordings per species would have further reduced the training or test sample size. The hyperparameters, learning rate and weight decay, were taken from the grid search outcomes of the first experiment. The models were trained for 50 epochs without validation-based early stopping and the checkpoint with the lowest training-loss was used for test-set evaluation.

\subsection{Experiment 3: existing pretrained models}
BatDetect2 and BAT~\citep{macaodhaGeneralApproachBat2022,fundel2023Automatic} were evaluated without additional training on both the natural-IPI and normalized-IPI datasets. Because the models produced outputs at different levels, their predictions were converted to a common recording-level format. For BatDetect2, the species occurring most frequently among the detections in a recording was taken as the recording-level prediction. For BAT, the highest-ranked species prediction was retained. Evaluation was based on these predicted labels rather than their associated probabilities, and a prediction was considered correct when it matched the recording's reference species label.

Because BatDetect2 and BAT had originally been trained on different sets of European bat species, the evaluation was conducted on two scopes. First, each model was evaluated on the complete dataset. Second, each model was evaluated only on recordings belonging to species represented in its original training data. The latter therefore constituted a model-specific known-species subset.

\subsection{Evaluation metrics}
Performance was assessed using accuracy and macro-averaged precision, recall, and F1-score. Macro-averaging assigned equal weight to each species and was used because the dataset remained imbalanced. Per-species precision, recall, and F1-score were also calculated to examine variation in performance among species.

For Experiment 1, metrics were calculated separately for each validation fold and summarized as mean and standard deviation across the five folds. For Experiments 2 and 3, metrics were calculated on the corresponding fixed evaluation datasets.
\section{Results}
A complete summary of classification performance and macro F1-scores across all three experimental conditions is provided in Table~\ref{tab:overall_results}.

\subsection{Dataset composition and IPI normalization}
Audio processing yielded two parallel datasets comprising 2-second segments of bat call sequences across 16 species: a natural IPI dataset preserving original temporal intervals and a normalized IPI dataset with inter-pulse intervals fixed to 50 ms. Each condition contained 251 segments (502 seconds total audio), with paired file naming ensuring identical track matching across conditions. For the within-condition cross-validation (Experiment 1), each dataset was partitioned using an 80/20 stratified train–validation split (201 training, 50 validation segments) for each of the 5-kfold splits, averaging the results. The cross-condition evaluation (Experiment 2) utilized an equivalent 80/20 train–test split (201 training, 50 testing segments), while the benchmark evaluation of existing classifiers (Experiment 3) evaluated all 251 segments per dataset condition.

\begin{figure*}
	\centering
		\includegraphics[width=\textwidth]{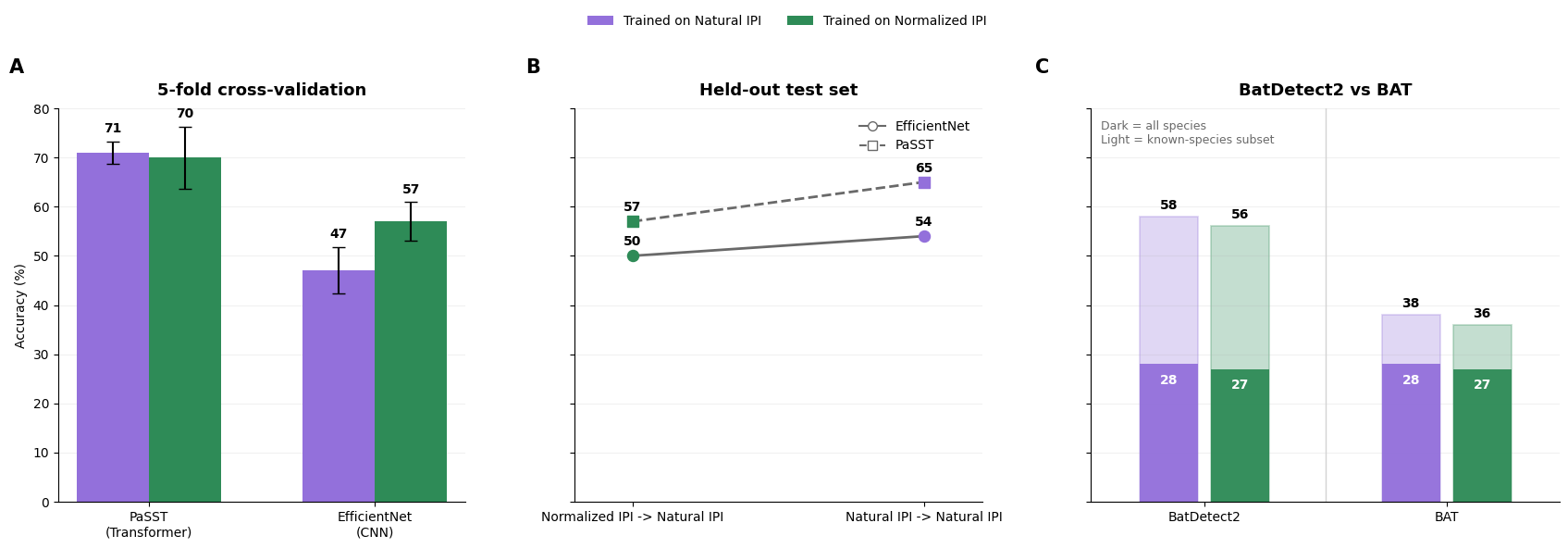}
	\caption{Classification accuracy across the three experiments. \textbf{(A)} Within-condition performance of EfficientNet-B0 and PaSST under the natural-IPI and normalized-IPI conditions. In each case, models were trained and validated on recordings from the same IPI condition using five-fold stratified cross-validation. \textbf{(B)} Cross- and matched-condition performance in Experiment 2. EfficientNet-B0 and PaSST were trained on normalized-IPI (left) and natural-IPI (right) recordings and evaluated on the same fixed test set with natural-IPI variation. Connected points show the change from training on normalized-IPI recordings (normalized $\rightarrow$ natural) to training on their natural-IPI counterparts (natural $\rightarrow$ natural). Experiment 2 was the only experiment in which models were evaluated under an IPI condition different from that used for training. \textbf{(C)} Performance of the existing pretrained classifiers BatDetect2 and BAT when evaluated without additional training on the natural-IPI and normalized-IPI datasets. Bars show results for the complete dataset and each model's known-species subset. }
	\label{fig:results}
\end{figure*}

\begin{figure*}
	\centering
		\includegraphics[width=\textwidth]{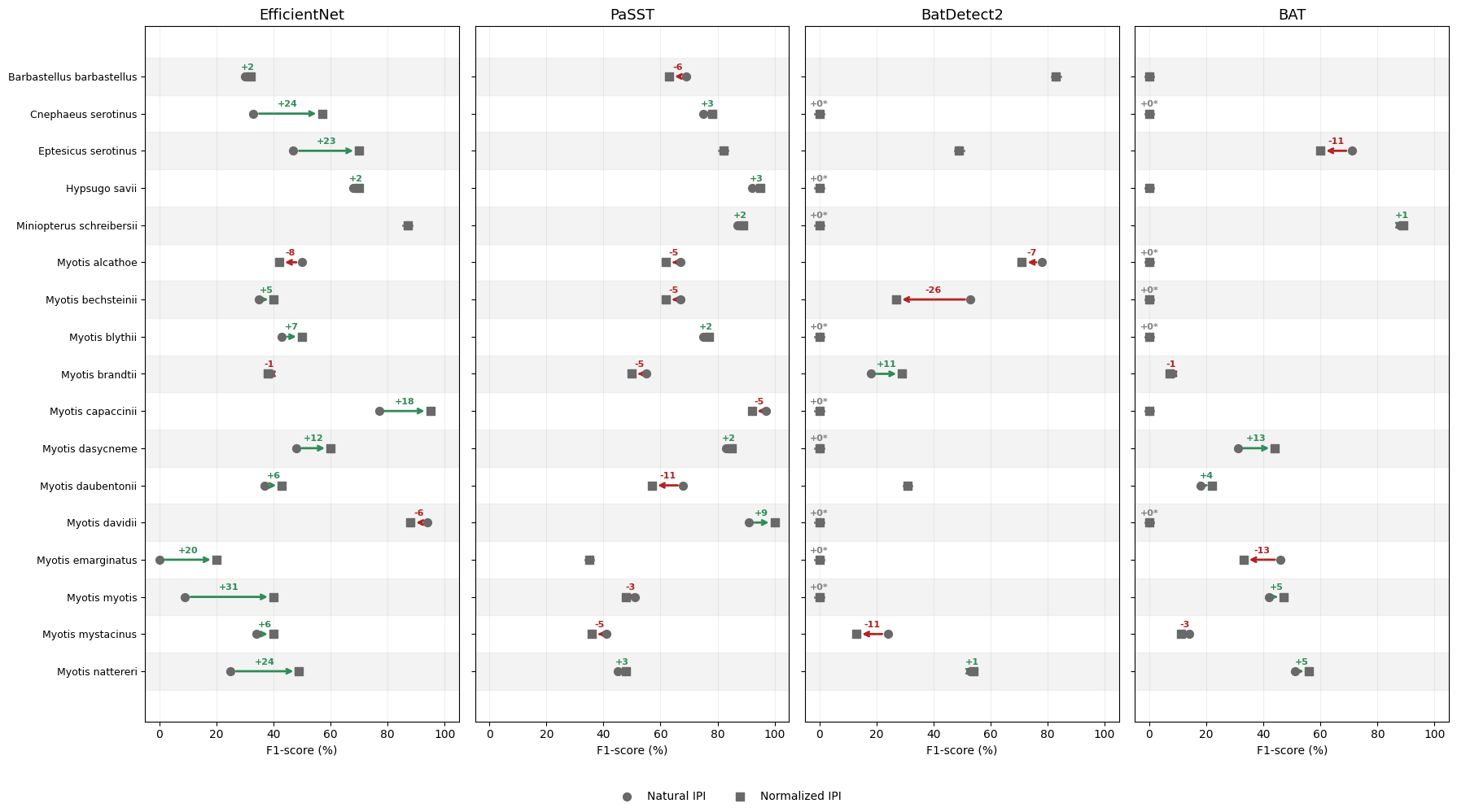}
	\caption{\textbf{Effect of IPI normalization on species-level F1-scores.} Species-level F1-scores under the natural-IPI and normalized-IPI conditions are shown for EfficientNet-B0, PaSST, BatDetect2, and BAT (from left to right). For EfficientNet-B0 and PaSST, scores were obtained when each model was trained and evaluated within the same IPI condition. BatDetect2 and BAT were evaluated without additional training on both dataset conditions. Arrows indicate the change from the natural-IPI score to the normalized-IPI score for each species. Green right-pointing arrows denote an increase following IPI normalization, whereas red left-pointing arrows denote a decrease. Numerical labels report the corresponding change in F1-score in percentage points, calculated as normalized IPI minus natural IPI. In the BatDetect2 and BAT panels, an asterisk indicates a species that was not represented in the respective model's original training data; asterisks do not indicate statistical significance.} 
	\label{fig:results_species}
\end{figure*}

\subsection{Experiment 1: within-condition model training and cross-validation}
 To determine whether temporal sequence context impacts classification performance, we compared model accuracies when trained and evaluated on natural versus normalized IPI datasets (Figure~\ref{fig:results}A). Normalizing IPI minimally affected PaSST performance, causing a minor accuracy reduction from 71\% ± 2.3\% on natural IPI to 70\% ± 6.3\% on normalized IPI. In contrast, EfficientNet showed a 10\% performance gain under the normalized condition (57\% ± 3.9\%) relative to the natural condition (47\% ± 4.7\%). Overall, PaSST consistently outperformed EfficientNet across both dataset conditions. Species-level analysis (Figure~\ref{fig:results_species}) revealed that PaSST maintained moderate F1 scores even for under-represented species (e.g. \textit{Myotis blythii}, \textit{M. alcathoe}, and \textit{M. bechsteinii}). EfficientNet exhibited greater performance fluctuation across species compared to PaSST.
 
\subsection{Experiment 2: cross-condition held-out evaluation}
To evaluate model generalizability to realistic audio, models trained on normalized IPI data were evaluated on a held-out test set containing natural IPI variation (Figure~\ref{fig:results}B). Evaluating models on mismatched temporal contexts caused a moderate performance drop across both architectures compared to the matched baseline where the models were both trained and evaluated on natural IPI variation. PaSST accuracy decreased by 8 percentage points (from 65\% to 57\%), while EfficientNet accuracy decreased by 4 percentage points (from 54\% to 50\%). Despite this drop, PaSST maintained higher overall accuracy than EfficientNet, and both models retained reasonable classification capacity.

\subsection{Experiment 3: existing pretrained models}
To investigate whether normalizing IPI variation affects the accuracy of existing classifiers, BatDetect2 and BAT were evaluated on both datasets without retraining (Figures~\ref{fig:results}C and~\ref{fig:results_species}). Normalization had minimal effect. Overall performance was poor: both BatDetect2 and BAT achieved 28\% accuracy on the natural-IPI dataset and 27\% on the normalized-IPI dataset.

Restricting evaluation to species represented in each model's training taxonomy increased performance, while the effect of IPI normalization remained small. BatDetect2 achieved 58\% accuracy on the natural-IPI dataset and 56\% on the normalized-IPI dataset, a decrease of two percentage points. BAT achieved 38\% and 36\%, respectively.

\begin{table*}[htbp]
\centering
\caption{Classification performance across experimental conditions for custom-trained models and pretrained benchmark classifiers. Values represent accuracy or macro F1-score, as indicated. For Experiment 1, values are reported as the mean ($\pm$ SD) across the five validation folds. Experiments 2 and 3 used single evaluation sets and are therefore reported without variability estimates.}
\label{tab:overall_results}
\small
\begin{tabular}{llcc}
\toprule
\textbf{Experiment / Model} & \textbf{Dataset Condition} & \textbf{Accuracy (\%)} & \textbf{Macro F1-Score} \\
\midrule
\multicolumn{4}{l}{\textbf{Experiment 1: Within-Condition Cross-Validation}} \\
\hspace{1em} PaSST & Natural IPI  & $71.0 \pm 2.3$ & 0.68 \\
\hspace{1em} PaSST & Normalized IPI & $70.0 \pm 6.3$ & 0.67 \\
\hspace{1em} EfficientNet-B0 & Natural IPI & $47.0 \pm 4.7$ & 0.43 \\
\hspace{1em} EfficientNet-B0 & Normalized IPI & $57.0 \pm 3.9$ & 0.52 \\
\midrule
\multicolumn{4}{l}{\textbf{Experiment 2: Cross-Condition Generalization}} \\
\hspace{1em} PaSST & Natural Train $\rightarrow$ Natural Test & 65.0 & 0.60 \\
\hspace{1em} PaSST & Normalized Train $\rightarrow$ Natural Test & 57.0 & 0.54 \\
\hspace{1em} EfficientNet-B0 & Natural Train $\rightarrow$ Natural Test & 54.0 & 0.52 \\
\hspace{1em} EfficientNet-B0 & Normalized Train $\rightarrow$ Natural Test & 50.0 & 0.46 \\
\midrule
\multicolumn{4}{l}{\textbf{Experiment 3: Pretrained Classifier Benchmarks}} \\
\hspace{1em} BatDetect2 (All Species) & Natural IPI & 28.0 & 0.18 \\
\hspace{1em} BatDetect2 (All Species) & Normalized IPI & 27.0 & 0.16 \\
\hspace{1em} BatDetect2 (Known Species) & Natural IPI & 58.0 & 0.41 \\
\hspace{1em} BatDetect2 (Known Species) & Normalized IPI & 56.0 & 0.38 \\
\hspace{1em} BAT (All Species) & Natural IPI & 28.0 & 0.15 \\
\hspace{1em} BAT (All Species) & Normalized IPI & 27.0 & 0.16 \\
\hspace{1em} BAT (Known Species) & Natural IPI & 38.0 & 0.24 \\
\hspace{1em} BAT (Known Species) & Normalized IPI & 36.0 & 0.29 \\
\bottomrule
\end{tabular}
\end{table*}

\section{Discussion}

This study investigated whether variation in interpulse intervals contributes to automated bat-species classification and whether transformer-based models use this information more effectively than CNN-based models. Across the three experiments, we found limited evidence that the investigated classifiers relied strongly on natural IPI variation. When EfficientNet-B0 and PaSST were trained and evaluated within the same IPI condition (Experiment 1), PaSST performance remained similar across conditions, whereas EfficientNet-B0 performed better on recordings with normalized IPI. In the cross-condition evaluation (Experiment 2), however, performance decreased when models trained on normalized-IPI recordings were evaluated on recordings retaining natural temporal variation. Finally, the existing classifiers BatDetect2 and BAT showed only small performance differences between natural- and normalized-IPI recordings (Experiment 3). Together, these findings provide limited evidence that IPI variation was a dominant source of species-discriminative information under the conditions examined. They also indicate that performance gains obtained on temporally normalized recordings may not transfer directly to natural recordings.

\subsection{Sensitivity to IPI variation}
The limited differences observed for BatDetect2 and BAT indicate that normalizing IPI had relatively little effect on their recording-level predictions. This result does not demonstrate that the models ignored temporal information altogether. Instead, it suggests that the specific variation removed by the normalization procedure contributed little to their predictions relative to other available acoustic features. Spectral and morphological properties of individual calls, including frequency, bandwidth, duration, and modulation structure, may have provided sufficient information for many classifications. Nevertheless, determining the relative contribution of these features would require methods designed to attribute model predictions to specific parts of the input.

The response of the fine-tuned models provides further context. Contrary to the expectation that reducing IPI variation would affect PaSST more strongly than EfficientNet, both models achieved higher within-condition performance on the normalized-IPI dataset. The improvement was larger for EfficientNet. One possible explanation is that IPI normalization reduced temporal variability that was unrelated to species identity, thereby simplifying the classification problem. Under this interpretation, the models could learn more consistent associations between call morphology and species labels when the temporal arrangement of calls was normalized.

Other explanations cannot be excluded. The normalization procedure inserted or removed background audio to obtain the target intervals and may therefore have changed acoustic characteristics beyond IPI. These modifications could have introduced discontinuities or repeated background patterns that made the normalized recordings easier to classify. Manual onset annotation also introduced uncertainty in the positions of the true acoustic onsets. The study should therefore be understood as comparing natural recordings with recordings subjected to the complete normalization procedure, rather than as a perfectly isolated manipulation of IPI. Moreover, because the dataset was relatively small and remained imbalanced, the cross-validation estimates may be sensitive to the allocation of individual recordings across folds. The improved within-condition performance consequently does not establish that normalized recordings contained more biologically meaningful species information.

\subsection{Generalization to natural recordings}
The held-out experiment provided further context for the within-condition results. Although IPI normalization improved EfficientNet performance when training and evaluation were conducted under the same condition, this advantage did not transfer to natural recordings. For both architectures, models trained on natural-IPI recordings performed better on the fixed natural-IPI test set than models trained on normalized-IPI recordings. This suggests that training on naturally varying recordings better prepared the models for the target evaluation condition.

This finding should not be interpreted as evidence that the models necessarily learned IPI as a species-specific feature. Exposure to natural-IPI recordings may instead help a model accommodate the broader variability encountered in realistic acoustic data, including variation that is not directly informative of species identity. Conversely, normalization may improve performance when training and evaluation data are processed identically while reducing robustness to unmodified recordings. This distinction is relevant to passive acoustic monitoring, for which models must ultimately operate on naturally varying field recordings.

The cross-condition results should nevertheless be interpreted in light of the limited test-set size. Each species was represented by approximately four test recordings, and no separate validation set was available. Although the matched natural-IPI and normalized-IPI test sets contained the same source recordings, estimates of the performance decrease may therefore be sensitive to the particular recordings included in the test partition. The procedures used to select hyperparameters and determine the training duration should also be considered when interpreting these results.

More broadly, the source recordings were weakly labelled and may contain uncertainty in species identity, variation in recording quality, or calls from non-target species. The two-second clips also capture only a limited portion of the behavioural and environmental variation encountered in passive acoustic monitoring. The extent to which these findings generalize to other species, regions, recording devices, and field conditions therefore remains to be established using larger and more diverse datasets.

An additional limitation concerns the taxonomic labels of the serotine recordings. \textit{Cnephaeus serotinus} and its former combination \textit{Eptesicus serotinus} were inadvertently treated as separate target classes during model training and evaluation. The models were therefore required to distinguish two labels representing the same biological species and may have learned differences associated with recording source, location, or recording conditions rather than species identity. Species-level results for these two labels should consequently not be interpreted as biological discrimination, and their duplication may also have affected aggregate metrics, particularly macro-averaged scores, by assigning additional weight to this taxon. Because both labels occurred under the natural- and normalized-IPI conditions, the comparison between IPI conditions may be less affected than the absolute performance estimates; nevertheless, the resulting comparisons should be interpreted cautiously.

\subsection{Comparison of model pipelines}
PaSST achieved higher performance than EfficientNet across the fine-tuning experiments, suggesting that PaSST is a promising model for classifying the species included in this dataset. This difference cannot, however, be attributed specifically to its transformer architecture or to a greater ability to exploit temporal context. EfficientNet and PaSST differed not only in architecture but also in their pretraining datasets, input representations, spectrogram parameters, temporal resolutions, and preprocessing procedures. Their comparison therefore concerns two complete modelling pipelines rather than an isolated comparison between convolution and self-attention.

The results from BatDetect2 and BAT likewise do not provide a controlled architectural comparison. BatDetect2 performed better than BAT in the present evaluation, but the models were developed using different training datasets, species sets, output formats, and prediction procedures. The recording-level predictions were also derived differently: BatDetect2 predictions were aggregated across detections, whereas BAT predictions were based on the highest-ranked species. Furthermore, the model-specific known-species evaluations contained different subsets of species. The combined results therefore leave open whether CNN- or transformer-based architectures are generally better suited to bat-species classification.

The pretrained classifiers used the former label \textit{Eptesicus serotinus}. Because this name refers to the same taxon as \textit{Cnephaeus serotinus}, recordings carrying either name represent a species included in the models' original training taxonomy. Apparent errors between these two names reflect nomenclatural inconsistency rather than discrimination between biological species.

Importantly, PaSST's higher performance does not by itself show that it exploited temporal information other than IPI. Establishing this would require additional manipulations, such as changing call order while preserving IPI, comparing complete sequences with isolated calls, or masking spectral and temporal features independently.

\subsection{Implications}
Our results have two main implications for automated bat monitoring. First, natural variation in IPI did not appear to be a major determinant of classification performance for the models and recordings examined. This does not imply that temporal context is irrelevant to bat classification more generally. Temporal information includes call order, changes in call morphology across a sequence, rhythm, and longer-term behavioural patterns, none of which was independently manipulated in this study. The findings are therefore specific to variation in the spacing between consecutive call onsets within two-second recordings.

Second, the higher within-condition performance obtained after IPI normalization illustrates that making a dataset more regular can simplify classification without necessarily improving performance on natural recordings. Evaluations conducted only on similarly processed training and test data may consequently overestimate performance under field conditions. Training and evaluation datasets for passive acoustic monitoring should retain representative natural variability, and cross-condition tests can help identify models that depend on properties introduced by preprocessing.

Future studies could extend this framework using larger datasets, additional species, and recordings collected across multiple locations and devices. More targeted ablation experiments could independently manipulate IPI, call order, call morphology, and background structure. Evaluating natural- and normalized-IPI training conditions on the same untouched natural test set would provide a direct measure of how temporal normalization affects generalization. Comparisons between CNNs and transformers would also be strengthened by matching their input representations, pretraining conditions, and optimization procedures.

By experimentally modifying call timing, this study provides a framework for examining which aspects of bat-call sequences contribute to automated classification. Such analyses can help move acoustic classifiers beyond benchmark performance alone and towards a clearer understanding of when their predictions remain reliable under the natural variability encountered in large-scale biodiversity monitoring.

\section{Conclusion}
This study investigated whether variation in interpulse intervals contributes to automated bat-species classification and whether transformer-based models are more sensitive to this information than CNN-based models. The results provide limited evidence that natural IPI variation was a dominant source of species-discriminative information under the conditions examined. BatDetect2 and BAT showed only small differences between natural- and normalized-IPI recordings, while reducing IPI variation did not affect PaSST more strongly than EfficientNet. Instead, within-condition performance remained similar for PaSST and improved for EfficientNet following IPI normalization.

The cross-condition evaluation nevertheless showed that performance obtained under normalized temporal conditions did not transfer fully to natural recordings. When models trained on normalized-IPI recordings were evaluated on matched recordings with natural IPI variation, accuracy decreased for both architectures. This finding indicates sensitivity to differences between the training and evaluation conditions, although it does not establish whether the decrease resulted specifically from IPI variation or from other changes introduced by the normalization procedure.

Overall, the findings suggest that normalizing call spacing can simplify within-condition classification without necessarily improving performance on unmodified recordings. They also demonstrate the importance of evaluating acoustic classifiers on data that retain the natural variability expected during deployment. Larger and more diverse datasets, together with more targeted manipulations of temporal and spectral information, will be needed to determine how sequence context contributes to bat-species classification and whether transformer-based models use it differently from CNN-based models.


\bibliographystyle{cas-model2-names}
\bibliography{IPI_main}

\printcredits

\section*{Code availability}
The code used for the annotation tool training and evaluation is publicly available at: https://github.com/welmoed505/thesis-bat-sound-classifiers.git 

\section*{Data availability}
The source recordings used in this study are available through the ChiroSetEurope dataset~\citep{baier2026CSEpaper} at: https://doi.org/10.5281/zenodo.20773227

\section*{Funding}
This work was supported by funding received via the Horizon Europe projects TETTRIs (grant no. 101081903, supporting work by BG) and MAMBO (grant no. 101060639, supporting work by ALB, BG, and DS). The views and opinions expressed are, however, those of the authors only and do not necessarily reflect those of the European Union or the European Research Executive Agency (REA). The funders had no role in study design, data collection and analysis, decision to publish, or preparation of the manuscript.


\end{document}